\documentclass[sigconf,screen]{acmart}

\AtBeginDocument{%
  }

\usepackage[capitalize,noabbrev]{cleveref}
\usepackage{colortbl}
\usepackage{arydshln}
\acmConference[PoliSim@CHI 2026]{PoliSim@CHI 2026: LLM Agent Simulation for Policy}{April 16, 2026}{Barcelona, Spain}
\renewcommand{\footnotetextcopyrightpermission}[1]{%
  \footnotetext{%
    This paper was prepared for \href{https://polisim.net/}{PoliSim@CHI 2026: LLM Agent Simulation for Policy}, a workshop at \href{https://chi2026.acm.org/}{CHI 2026 (CHI Conference on Human Factors in Computing Systems)}, April 16, 2026, Barcelona, Spain.
  }%
}

\begin{document}

\title[Counterfactual Semantics for Policy Evaluation in Simulated Online Communities]{From Plausible to Causal: Counterfactual Semantics for Policy Evaluation in Simulated Online Communities}

\author{Agam Goyal}
\authornote{Both authors contributed equally.}
\orcid{0009-0009-5989-2887}
\affiliation{%
  \institution{University of Illinois Urbana-Champaign}
  \city{Urbana}
  \state{Illinois}
  \country{USA}}
\email{agamg2@illinois.edu}

\author{Yian Wang}
\authornotemark[1]
\orcid{0000-0001-9545-3499}
\affiliation{%
  \institution{University of Illinois Urbana-Champaign}
  \city{Urbana}
  \state{Illinois}
  \country{USA}}
\email{yian3@illinois.edu}

\author{Eshwar Chandrasekharan}
\authornote{Both authors are advisors of this work.}
\orcid{0000-0002-7473-1418}
\affiliation{%
  \institution{University of Illinois Urbana-Champaign}
  \city{Urbana}
  \state{Illinois}
  \country{USA}}
\email{eshwar@illinois.edu}

\author{Hari Sundaram}
\authornotemark[2]
\orcid{0000-0003-3315-6055}
\affiliation{%
  \institution{University of Illinois Urbana-Champaign}
  \city{Urbana}
  \state{Illinois}
  \country{USA}}
\email{hs1@illinois.edu}

\renewcommand{\shortauthors}{Goyal et al.}

\begin{abstract}
LLM-based social simulations can generate believable community interactions, enabling ``policy wind tunnels'' where governance interventions are tested before deployment. But believability is not causality. Claims like ``intervention $A$ reduces escalation'' require causal semantics that current simulation work typically does not specify. We propose adopting the causal counterfactual framework, distinguishing \textit{necessary causation} (would the outcome have occurred without the intervention?) from \textit{sufficient causation} (does the intervention reliably produce the outcome?). This distinction maps onto different stakeholder needs: moderators diagnosing incidents require evidence about necessity, while platform designers choosing policies require evidence about sufficiency. We formalize this mapping, show how simulation design can support estimation under explicit assumptions, and argue that the resulting quantities should be interpreted as simulator-conditional causal estimates whose policy relevance depends on simulator fidelity. Establishing this framework now is essential: it helps define what adequate fidelity means and moves the field from simulations that look realistic toward simulations that can support policy changes.
\end{abstract}



\maketitle

\section{The Causal Gap in Simulation-Based Governance of Online Communities}

Online communities today are not merely social spaces, but rather they are governance systems~\cite{weld2025perceptions}. Every moderation action or rule enactment on platforms such as Reddit and X constitutes a policy decision with consequences for millions of users~\cite{kraut2012building}.  These policy decisions have widespread consequences such as community norm~\cite{10.1145/3134666}, linguistic dynamics~\cite{goyal2025language,10.1145/3613904.3642769}, newcomer retention~\cite{russo2025does}, and public perception of the platform. Despite these stakes, community governance remains largely reactive~\cite{scheuerman2021framework}, where platform designers and moderators look at outcomes of policy changes and interventions in a post-hoc manner which forecloses a huge opportunity for \textit{proactive governance} which can help in anticipating how interventions might reshape community dynamics before deployment~\cite{lambert2024positive,lambert2025mind}.

The HCI research community has previously identified the lack of prototyping techniques in social computing as a key reason for this gap for this gap~\cite{park2022social}, highlighting that designers lack robust ways to proactively reflect on the downstream consequences that may emerge from such interventions~\cite{schon2017reflective}. As a result, the paradigm of LLM-based social simulations has emerged as a lightweight prototyping technique to test design choices and their potential downstream consequences~\cite{park2023generative}. However, in this paper we argue that currently, these simulation-based prototyping techniques can help us map a set of \textit{plausible} outcomes, but cannot yet answer \textit{causal} questions---whether deployment of intervention A will reliably cause outcome B. 

To see why this distinction matters, consider the kinds of questions that governance stakeholders need to answer. A moderator reviewing a thread that escalated to a toxic outcome would ask \emph{``Was the initial inflammatory comment the cause of this outcome? Would the thread have stayed civil without it?''} A platform designer evaluating an automated counter-speech system would ask \emph{``If we deploy this at scale, will it reliably prevent escalation?''} Finally, a policy team weighing rule changes would ask \emph{``How much of the toxicity reduction we observed was due to our intervention versus an independent shift in community norms?''} These are causal questions, and none of them can be answered by demonstrating that a simulation produces \textit{believable} content.

We argue that, for simulations to inform governance decisions, the field must move from asking ``Do these simulations look real?'' to asking ``What causal claims can these simulations support?'' This is not a critique of existing work: progress on behavioral plausibility has been essential and foundational. Rather, we see this as the next conceptual step. Governance stakeholders do not only need evidence that an intervention changes outcomes on average; they need to know whether a specific factor was necessary for a harmful outcome, whether an intervention is sufficient to reliably prevent it, and under what assumptions such conclusions hold. If simulations are to inform these decisions, the evidence they produce must be legible in causal terms, for example, ``the probability that the outcome would not have occurred without the intervention is $X$.'' 

To this end, we propose adopting the causal counterfactual framework for event attribution developed by \citet{hannart2016causal} to analyze simulated interventions in online communities. Its key advantage is that it distinguishes two different causal questions that governance stakeholders routinely conflate: whether a factor was necessary for a harmful outcome, and whether an intervention is sufficient to reliably produce a desired one. This distinction enables simulations to support precise, probabilistic, and policy-relevant claims from model outputs, rather than only informal judgments that an intervention ``seemed to help.'' The framework has been influential in fields such as climate-event attribution, and we argue that it provides the missing causal semantics for simulation-based policy evaluation in social computing.

The remainder of this paper develops our central position. Section~\ref{sec:related_works} presents an overview of the literature landscape on LLM simulations. Section~\ref{sec:framework} introduces the causal counterfactual framework, grounding its key concepts in concrete governance examples. Section~\ref{sec:mapping} maps the framework onto simulation-based policy evaluation, showing how different types of causal evidence serve different governance stakeholders. Section~\ref{sec:fidelity} presents an important pre-requisite about simulator fidelity under this framework. Finally, Section~\ref{sec:agenda} outlines a research agenda for operationalizing causal claims in simulated communities.
\section{Related Works}~\label{sec:related_works}

\noindent Recent work on LLM-based social simulation has demonstrated that language-model agents can generate plausible social interactions and support early-stage design exploration for social computing systems. For social computing specifically, \citet{park2022social} introduced the idea of populated prototypes, using language models to help designers reason about social interactions before deployment, and showed how generative agents can produce believable, persistent, and socially situated behavior in interactive environments~\cite{park2023generative}. More recent work has pushed this paradigm toward larger-scale and more empirically grounded simulation, including efforts to simulate populations and known human-group dynamics at scale~\cite{park2024generative,yang2024oasis,chuang2023wisdom,chuang2024beyond,chuang2024simulating}. These works overall argue that LLM social simulations may become a useful research method under appropriate validation and scope conditions~\cite{anthis2025llm}. Domain-specific work has also begun to study content dissemination and regulation in multi-agent social media simulation settings~\cite{liu-etal-2025-mosaic,cai2025simulation} including agent-native social networks such as Moltbook~\cite{feng2026moltnet,goyal2026social}. However, this literature has primarily focused on plausibility, human-behavioral resemblance, or methodological promise, rather than on the semantics of what kinds of causal claims simulated interventions can justify. At the same time, the online community governance literature has established that real interventions such as bans, quarantines, moderation explanations, and visible norm-setting can meaningfully reshape participation and discourse~\cite{10.1145/3274301,10.1145/3134666,jhaver2019does,matias2019preventing,chandrasekharan2022quarantined,goyal2024uncovering,weld2025perceptions,goyal2025language}. This line of work provides both concrete intervention classes and real-world outcome measures against which simulation-based policy evaluation could eventually be calibrated. 

Our contribution is complementary to both strands, and our central position is that policy-facing simulation requires an explicit causal semantics. To that end, we adapt the causal counterfactual framework of necessary and sufficient causation~\cite{pearl2009causality,hannart2016causal}, and position probabilities of necessity and sufficiency as a way to connect simulated intervention outcomes to the distinct attribution and decision needs of moderators, platform designers, and policy teams.
\section{Causal Counterfactual Theory for Community Governance}
\label{sec:framework}

We now introduce the causal machinery our framework relies on. We keep the treatment deliberately concise as our goal is to equip the reader with the specific concepts needed to follow our arguments in Section~\ref{sec:mapping}, rather than providing a detailed primer.

When we think about causality, the phrase ``$X$ caused $Y$'' remains ambiguous, as it can mean two distinct things, with the distinction being critical for governance applications:

\begin{itemize} 
\item \textbf{Necessary causation.} Had $X$ not occurred, $Y$ would not have occurred either. The \emph{Probability of Necessary Causation} (PN) quantifies this as $\text{PN} = P(Y_0 = 0 \mid Y = 1, X = 1)$. In words, given that the outcome occurred and the cause was present, what is the probability the outcome would \emph{not} have occurred  without the cause ($Y_0=0$)?
\item \textbf{Sufficient causation.} Introducing $X$ reliably produces $Y$. The \emph{Probability of Sufficient Causation} (PS) quantifies this as $\text{PS} = P(Y_1 = 1 \mid Y = 0, X = 0)$.
In words, given that the outcome did \emph{not} occur and the cause was absent, what is the probability the outcome \emph{would} have occurred had the cause been introduced ($Y_1=1$)?
\end{itemize}

We will show in Section~\ref{sec:mapping} that this distinction maps directly onto the different information needs of different governance stakeholders.

\section{Mapping the Framework to Simulation-Based Community Governance} \label{sec:mapping} 

We now show how the causal counterfactual framework maps onto the concrete setting of simulation-based policy evaluation for online communities. We define the key variables, discuss the assumptions, show what PN and PS mean in governance terms, and identify which stakeholders need which type of causal evidence. 

\subsection{Defining the Variables} In the governance context, the variables in the causal framework take on specific meanings: 

\begin{itemize} 

\item \textbf{Intervention ($X$).} A governance action injected into the simulation. This could be a moderation action (removing a post, issuing a warning), a design change (altering reply visibility, adding friction to posting), a counter-speech strategy (automated empathetic response, community-norm reminder), or a rule change (new community guidelines, updated content policy). 

\item \textbf{Outcome ($Y$).} A measurable discourse property of the simulated thread or community. Examples include escalation to harassment ($Y = 1$ if toxicity exceeds a threshold), thread derailment, user departure from platform, or norm compliance. The outcome must be operationalized with a clear threshold

 \item \textbf{Factual and counterfactual conditions.} For each simulated scenario, the \emph{factual} condition is the simulation run as configured (with or without intervention), and the \emph{counterfactual} is the same scenario re-run with the intervention toggled. Because the simulation is computational, we can run both---something impossible in the real world, where every thread unfolds only once. \end{itemize}

 However, PN and PS as we defined above involve counterfactual quantities that are not directly observable. However, under two assumptions of the causal attribution framework, they reduce to simple functions of estimable probabilities:\\

 \noindent\textbf{Exogeneity:} Exogeneity implies that \emph{the cause $X$ is assigned independently of the system's dynamics, i.e., it is not influenced by confounders that also affect $Y$}. 
 
The assumption of exogeneity is one that randomized experiments satisfy by design. However, in observational studies of real platforms, interventions are not randomly assigned:, but are rather entangled with the very dynamics it aims to influence, making it difficult to disentangle cause from context. For example, communities may adopt stricter rules \emph{because} they experience more toxicity. However, simulation resolves this by construction. When a researcher injects an intervention into a simulated thread, the injection is controlled by the experimental protocol, and not generated by the simulated agents' dynamics. This is also one of the core advantages of simulation-based evaluation, in that it converts what would be a quasi-experimental design in the real world into a true experiment, at least within the simulated/counterfactual world.

 \noindent\textbf{Monotonicity:} Monotonicity states that the cause $X$ operates in a consistent direction, i.e., it cannot increase the probability of $Y$ for some units and decrease it for others. 
 
 Monotonicity as a construct is more substantive. For many use cases relevant to governance, this is defensible, as removing an inflammatory comment should not \emph{increase} the probability of escalation, or introducing a cooldown period should not \emph{increase} impulsive hostile responses. However, social systems are reflexive, and monotonicity can fail because a counter-speech intervention perceived as patronizing might \emph{increase} hostility in some thread contexts while decreasing it in others. Similarly, a new community rule might improve behavior among norm-abiding users while triggering purposeful defiance from other groups of users. 

\begin{table*}[ht]
\centering
\small
\sffamily
\caption{Mapping governance stakeholders to causal evidence types. Different stakeholders require different types of causal evidence (necessary vs.\ sufficient causation), and the relevant outcome threshold varies by decision context.}
\label{tab:stakeholders}
\resizebox{\textwidth}{!}{
\begin{tabular}{p{2.5cm} p{5cm} p{2cm} p{3.5cm}}
\rowcolor{gray!10}\textbf{Stakeholder} & \textbf{Governance Question} & \textbf{Primary Metric} & \textbf{Decision Supported} \\
\midrule
\rowcolor{blue!10}Moderator & Was this action the cause of the harm? & PN & Attribution, accountability, case review \\
Platform Designer & Will this intervention reliably work at scale? & PS & Policy selection, resource allocation \\
\rowcolor{blue!10}Policy/Legal Team & Was the platform's inaction responsible for the outcome? & PN & Liability assessment, compliance \\
Research Team & Which interventions have the strongest causal effects? & PN and PS across thresholds & Evidence synthesis, intervention ranking \\
\bottomrule
\end{tabular}}
\end{table*}

Whenever these two assumptions hold, our formulations of PN and PS simplify to: 
\begin{equation} 
\text{PN} = 1 - \frac{p_0}{p_1}, \qquad \text{PS} = 1 - \frac{1 - p_1}{1 - p_0} \label{eq:pnps_simple} 
\end{equation}
where $p_1 = P(Y = 1 \mid X = 1)$ is the outcome probability when the cause is present and $p_0 = P(Y = 1 \mid X = 0)$ is the outcome probability when it is absent. These are simply outcome rates under two conditions, which is exactly what paired simulation runs yield. 

However, when monotonicity is violated, the expressions in Equation~\ref{eq:pnps_simple} would become a \emph{lower bound} on the true PN and PS. We however note that importantly, the violation of monotonicity themselves could reveal that an intervention has heterogeneous effects, which we argue is a governance-relevant finding on its own. A simulation framework can in fact detect these violations by examining the distribution of intervention effects across threads rather than relying solely on aggregate rates.

\subsection{What PN and PS Mean for Governance} With the definitions and assumptions above, we can now state precisely what PN and PS tell governance stakeholders.\\ 

\noindent\textbf{PN: Diagnosing specific incidents.} Suppose a simulated community thread escalated to severe harassment. A moderator or a social computing researcher using simulation to inform moderation practice could ask: \emph{Was the initial toxic comment the cause?} Then, \textit{given that escalation occurred in the presence of the toxic comment, PN is the probability that the thread would \textbf{not} have escalated had the comment been absent.}

A high PN (e.g., 0.85) means the toxic comment was very likely necessary for the escalation, i.e., the thread was otherwise on a civil trajectory, and the comment derailed it. A low PN (e.g., 0.20) means the thread was likely heading toward escalation regardless---the comment was incidental, not causal. This type of evidence supports attribution and accountability, enabling us to understanding which actions actually drive harmful outcomes, and by extension, where moderation effort is most impactful.\\

\noindent\textbf{PS: Evaluating policy interventions.} Now suppose a platform designer is considering whether the absence of counter-speech creates meaningful risk of escalation. Let $X=1$ denote the absence of counter-speech and $X=0$ denote the presence of counter-speech. Let the outcome $Y=1$ indicate that the thread escalates to severe harassment. Then, given that escalation did not occur when counter-speech was present, PS asks: \textit{what is the probability that the thread would have escalated had counter-speech been absent?}

A high PS (e.g., 0.80) means counter-speech is reliably effective, i.e., most threads that stayed civil did so \emph{because} of the intervention. A low PS (e.g., 0.20) means the civility was largely attributable to other factors (participant disposition, community norms, topic), and counter-speech was not the decisive factor. This type of evidence supports \textbf{policy selection and resource allocation}: deciding which interventions are worth deploying at scale and where the marginal value of investment is highest.\\

\noindent\textbf{A worked example:} To make this concrete, consider a simulation study that runs 1{,}000 threads under two conditions—with and without an automated counter-speech bot that responds to the first uncivil message. The outcome is defined as whether the thread escalates to severe harassment. Using the notation above, let $X=1$ denote the absence of counter-speech and $X=0$ denote its presence. Suppose the researchers observe:
\[
p_1 = P(Y=1 \mid X=1) = 0.30, \qquad
p_0 = P(Y=1 \mid X=0) = 0.08
\]

Computing PN and PS:
\[
PN = 1 - \frac{p_0}{p_1} = 1 - \frac{0.08}{0.30} = 0.73,
\qquad
PS = 1 - \frac{1-p_1}{1-p_0} = 1 - \frac{0.70}{0.92} = 0.24
\]

Here, $PN=0.73$ means that in threads that escalated when counter-speech was absent, there is a 73\% probability that the escalation would not have occurred had counter-speech been present. In this sense, the absence of counter-speech was likely necessary for the harmful outcome. 

By contrast, $PS=0.24$ means that among threads that remained civil when counter-speech was present, only 24\% would have escalated had counter-speech been absent. Thus, while counter-speech may matter substantially in preventing particular escalations, it is not by itself sufficient to explain most instances of civility.
\medskip

This asymmetry is governance-relevant. It suggests that counter-speech may be valuable as a targeted intervention for threads showing early warning signs, while also indicating that platform-wide civility cannot be attributed to counter-speech alone and likely depends on broader contextual and structural factors.

Note that a critical feature of this framework is that the causal evidence depends on how the outcome event $Y$ is defined---specifically, on where the thresholds are set. The same intervention can appear necessary for extreme outcomes but not sufficient for common ones, or vice versa. This is however not a limitation. Different governance decisions require different outcome definitions, and a safety team investigating a specific severe incident needs PN at an extreme threshold, while a product team deciding whether to invest in a counter-speech system needs PS at a moderate threshold. Reporting PN and PS across a range of outcome definitions provides a \emph{causal profile} of the intervention, which we argue is a richer characterization than any single effect size can offer.

Table~\ref{tab:stakeholders} summarizes how different governance stakeholders map onto the causal framework.

\section{Simulator Fidelity: A Prerequisite}\label{sec:fidelity} 

The validity of any PN or PS estimate depends on the fidelity of the simulator that produces it. If the simulated outcome rates $p_0$ and $p_1$ do not approximate the real-world quantities they are intended to represent, then the resulting causal claims are valid only for the simulated world rather than the governance setting of interest. Accordingly, the framework we propose should be understood as yielding simulator-conditional causal estimates: quantities that are meaningful and comparable within a simulation, but whose policy relevance depends on how well the simulator preserves the intervention-response structure of the real community context.

But what would it mean for a simulator to be ``good enough'' for governance? We argue that the framework provides a concrete answer: a simulator is adequate if its $p_0$ and $p_1$ approximate real-world values for the interventions and outcomes of interest. We also believe that this framework enforces epistemic honesty, as every claim must be qualified as ``in simulation, PN $= X$,'' making the gap between simulation and reality visible rather than obscured by vague assertions that ``the intervention seemed to help.'' The causal framework is what connects these existing empirical estimates to the outputs of simulation-based policy evaluation. 
\section{A Research Agenda} \label{sec:agenda} 

Adopting the causal counterfactual framework opens a concrete research program for moving simulation-based governance from plausible scenario generation toward policy-relevant causal evaluation. We organize this agenda around four priorities. Together, they address a progression from measurement, to calibration, to robustness, to practical use.

\subsection{Operationalization and Tooling} 

The first priority is to make PN/PS-style analysis operational on top of existing simulation platforms. Prior work in LLM-based social simulation has already produced reusable environments for populated prototyping, generative interaction, and large-scale agent-based social behavior. However, what is missing is a causal evaluation layer that sits on top of these systems. Concretely, this would include standardized protocols for paired factual/counterfactual runs from the same initial thread state, sufficient replications to estimate $p_0$ and $p_1$ with uncertainty, and reporting utilities that expose PN and PS across multiple outcome thresholds rather than only a single aggregate effect.

This tooling agenda matters because many governance decisions are threshold-dependent: an intervention may be highly relevant for preventing severe escalation, but much less informative for explaining ordinary civility. The immediate goal, then, is not merely to ``compute PN and PS,'' but to build causal profiling workflows that let researchers inspect how conclusions vary across thresholds, interventions, and thread contexts.

\subsection{Empirical Validation Against Real-World Estimates} 

The second priority is empirical calibration, which involves testing whether simulation-derived causal quantities track intervention effects observed in real online communities. This step is motivated by a large governance literature showing that real interventions such as bans~\cite{10.1145/3134666}, quarantines~\cite{chandrasekharan2022quarantined}, removal explanations~\cite{jhaver2019explanations}, and norm-setting~\cite{matias2019preventing} measurably alter participation and discourse. These studies provide intervention classes and outcomes for which there is already credible empirical evidence.

Accordingly, a natural validation strategy is to instantiate these same interventions in simulation and compare the resulting $p_0$, $p_1$, PN, and PS profiles against corresponding real-world estimates. Absolute agreement may be unrealistic, but even weaker forms of calibration would be informative. For example, can simulations preserve the rank-ordering of which interventions have larger versus smaller effects? Do they recover the same qualitative asymmetries across mild versus severe outcomes? Systematically documenting where simulations overestimate, underestimate, or mis-rank intervention effects would not only test fidelity, but also reveal which components of the simulator most need improvement. This validation agenda is also relevant for emerging AI-assisted governance systems. For example, recent work on community-specific content moderation suggests that language-model systems can be used to support operational decisions in online communities~\cite{kumar2024watch,zhan2025slm,goyal2025momoe,koshy2025venire,goyal2026vastu}. If such systems are to incorporate simulated evidence, then calibrating simulation-derived causal quantities against real intervention effects becomes even more important.

\subsection{Assumption Testing and Heterogeneous Effects in Community Contexts}

The third priority is to study when the assumptions underlying PN/PS estimation hold, and when they fail. In our setting, exogeneity is often satisfied by experimental design inside the simulator, because the intervention is injected by the researcher rather than endogenously generated by the agents. Monotonicity, however, is a much stronger empirical claim. Governance interventions are often socially reflexive, where a warning may calm one thread but inflame another. For e.g., a norm reminder may increase compliance in one community but trigger defiance in a different one. Prior work on online governance already suggests that interventions can have uneven effects across norms, communities, and user types~\cite{lambert2025does,weld2025perceptions}.

A strong research direction could then be to develop diagnostics that identify where interventions backfire, for whom, and under what local conditions. Rather than relying only on aggregate rates, future simulation studies should examine distributions of individual- or thread-level intervention effects, characterize clusters of positive and negative response, and connect those patterns to community structure, topic, norm salience, or participant role.

\subsection{Stakeholder-Centered Design of Causal Outputs}

The final priority is to ensure that the outputs of causal simulation are interpretable and decision-relevant for the people who would actually use them. Prior HCI and governance work has repeatedly shown that moderation and policy decisions are not made from abstract metrics alone, but through situated workflows, local norms, and practical tradeoffs~\cite{kraut2012building}. A metric is useful only if stakeholders understand what question it answers, what assumptions it depends on, and what level of evidence is sufficient for action.

For this reason, future work should not stop at estimating PN and PS, but also study how these quantities should be communicated, thresholded, and embedded into governance workflows. This could involve co-designing interfaces and reports with moderators, platform designers, and policy teams, and testing whether users correctly distinguish necessity-style claims from sufficiency-style claims when making intervention decisions.

\section{Conclusion} \label{sec:conclusion} 

Simulation-based policy evaluation for online communities is a promising paradigm, but its conceptual foundations remain incomplete. Existing work has shown that simulations can be believable. However, the next step is to ensure that they can support causal claims in order to be useful for policy implications. The causal counterfactual framework---by distinguishing necessary from sufficient causation, making assumptions explicit, and quantifying uncertainty---provides the semantics required for this shift. It clarifies the conditions under which simulation-derived evidence should be trusted and outlines a research agenda for making such evidence policy-relevant. We call on the social computing and simulation communities to adopt causal semantics as the standard for policy-facing claims.

\bibliographystyle{ACM-Reference-Format}
\bibliography{references}

\appendix
\end{document}